\documentclass[final,3p]{elsarticle}

\usepackage{lineno}
\usepackage[colorlinks=true,breaklinks=true,pdftex]{hyperref}
\usepackage{amsmath}
\usepackage{amssymb}
\usepackage{booktabs}
\usepackage{algorithm}
\usepackage{algorithmic}
\usepackage{multirow}
\usepackage{float}
\usepackage{bm}
\usepackage{tabularx}
\usepackage{wrapfig,booktabs}
\usepackage{subfig}
\usepackage{caption}
\usepackage{lipsum}
\usepackage{lineno}
\usepackage{appendix}
\usepackage{xcolor}
\modulolinenumbers[1]

\journal{CAGD; Special Issue of GMP 2020}

\biboptions{authoryear}

\begin{document}

\begin{frontmatter}

\title{LRC-Net: Learning Discriminative Features on Point Clouds by Encoding Local Region Contexts}

\author[first]{Xinhai Liu} 
\ead{lxh17@mails.tsinghua.edu.cn} 

\author[last]{Zhizhong Han}
\ead{h312h@umd.edu}

\author[first]{Fangzhou Hong} 
\ead{hongfz16@mails.tsinghua.edu.cn}

\author[first]{Yu-Shen Liu\corref{cor}} 
\ead{liuyushen@tsinghua.edu.cn} 
\cortext[cor]{Corresponding author} 

\author[last]{Matthias Zwicker} 
\ead{zwicker@cs.umd.edu}

\address[first]{School of Software, BNRist, Tsinghua University, Beijing, China} 
\address[last]{Department of Computer Science, University of Maryland, College Park, USA}

\begin{abstract}
Learning discriminative feature directly on point clouds is still challenging in the understanding of 3D shapes.
Recent methods usually partition point clouds into local region sets, and then extract the local region features with fixed-size CNN or MLP, and finally aggregate all individual local features into a global feature using simple max pooling.
However, due to the irregularity and sparsity in sampled point clouds, it is hard to encode the fine-grained geometry of local regions and their spatial relationships when only using the fixed-size filters  and individual local feature integration, which limit the ability to learn discriminative features.
To address this issue, we present a novel Local-Region-Context Network (LRC-Net), to learn discriminative features on point clouds by encoding the fine-grained contexts inside and among local regions simultaneously.
LRC-Net consists of two main modules.
The first module, named \emph{intra-region context encoding}, is designed for capturing the geometric correlation inside each local region by novel variable-size convolution filter.
The second module, named \emph{inter-region context encoding}, is proposed for integrating the spatial relationships among local regions based on spatial similarity measures.
Experimental results show that LRC-Net is competitive with state-of-the-art methods in shape classification and shape segmentation applications.
\end{abstract}

\begin{keyword}
Point clouds \sep Local region \sep Geometric information \sep Context
\end{keyword}

\end{frontmatter}

\section{Introduction}
As an important type of 3D data which can be acquired conveniently by various 3D sensors, point cloud has been increasingly used in diverse real word applications including autonomous driving \cite{qi2017frustum,yi2019hierarchical}, 3D modeling \cite{golovinskiy2009shape,gao2017bimtag,Zhizhong2016,HanCyber17a,zhong2019surface,skrodzki2018directional,zheng2018rolling,gao2015query}, indoor navigation \cite{zhu2017target} and robotics \cite{rusu2008towards}.
Therefore, there is an emerging demand to learn discriminative features with deep neural networks for 3D shape understanding.

Unlike images, point cloud is not suitable for the traditional convolutional neural network (CNN) which often requires some fixed spatial distribution in the neighborhood of each pixel.
 To alleviate this issue, an alternative way is to rasterize the point cloud into regular voxel representations and then apply 3D CNNs \cite{zhou2017voxelnet}.
However, the performance of plain 3D CNNs is largely limited by the serious resolution loss and the fast-growing computational cost, due to the inherent sparsity of 3D shapes.
To overcome the shortcoming of 3D CNNs, PointNet \cite{qi2017pointnet} was proposed as a pioneering work which directly learns global features for 3D shapes from point sets.
However, PointNet learns the feature of each point individually, while omitting the important contextual information among points.

To solve above-mentioned problems, recent studies have attempted to encode the local region contexts of point clouds with various designed manners.
Specifically, there are two kinds of local region contexts, including the intra-region geometric context and the inter-region spatial context.
On the one hand, some methods concentrate on capturing the context of geometric correlations inside each local region.
For example, PointNet++ \cite{qi2017pointnet++} uses a sampling and grouping strategy to hierarchically extract features for local regions.
More recently, Point2Sequence \cite{liu2018point2sequence} learns the contextual information inside a local region with an attention-based sequence to sequence network.
On the other hand, several studies attempt to utilize the context of spatial distribution information among local regions.
For example, KD-Net \cite{klokov2017escape} builds a kd-tree to divide the point cloud into small leaf bins and then hierarchically extracts the point cloud feature from the leaves to root according to a fixed spatial partition.
KC-Net \cite{shen2018mining} uses a graph pooling operation which can partially utilize the spatial distribution information among local regions.
However, it is still hard for these methods to encode the fine-grained contexts inside and among local regions simultaneously, especially for the geometric correlation between different scale areas inside a local region and the spatial relationships among local regions.
This motivates us to employ variable-size filters inside each local region and spatial similarity measures among local regions for capturing intra-region context information and inter-region context information, respectively.
Our method reliefs the limitation of the traditional CNNs in encoding the geometric context information on point clouds,  which usually implements a convolution layer with fixed-size filters, while the concrete filter size is a hyper-parameter.
To address above-mentioned problems, we propose LRC-Net to learn discriminative features from point clouds.

\begin{figure}[tp]
    \centering
    \includegraphics[width=10cm,height=6cm]{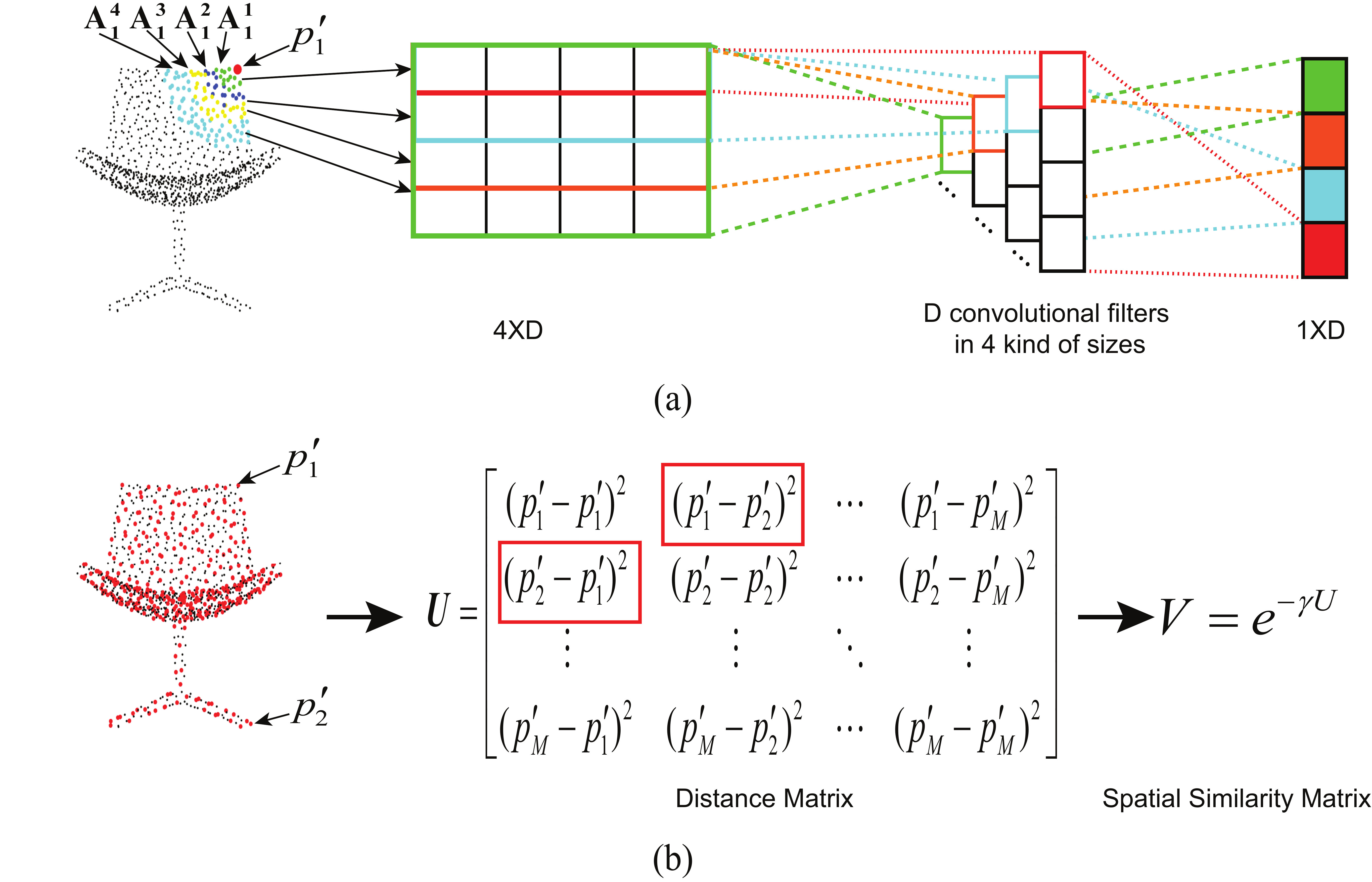}
    \caption{(a) \emph{intra-region context encoding}. It shows the process of capturing the geometric correlation inside the local region around the point $p'_1$.
(b) \emph{inter-region context encoding}. The spatial similarities of local regions are measured with the matrix $\bm{V}$, where two central points $p'_1$ and $p'_2$ of local regions are highlighted.}
    \label{fig:idea}
\end{figure}

Our key contributions are summarized as follows.
\begin{itemize}
  \item LRC-Net is presented for learning discriminative features directly from point clouds by simutaneously encoding the geometric correlation inside each local region and the spatial  relationships among local regions.
  \item \emph{Intra-region context encoding} module is designed for capturing the geometric correlation inside each local region by novel variable-size convolution filters, which learns the intrinsic structure and correction of multi-scale areas from their feature maps, rather than simple feature concatenating or max pooling as usually used in previous methods such as \cite{qi2017pointnet}.
  \item \emph{Inter-region context encoding} module is proposed for integrating the spatial relationships among local regions based on spatial similarity measures, which encodes the spatial distribution of local regions in their metric space.
\end{itemize}

The above two modules are illustrated in Figure \ref{fig:idea}.

\section{Related Work}
\textbf{Feature Learning from Regularized 3D Data.} Traditional methods \cite{liu2009robust,liu2009computing,liu2011computing,gao2015query,fehr2016covariance,zou2018broph,srivastava2019deeppoint3d,beksi2019topology,zhao2020hoppf} focus on cpturing the geometric information of 3D shapes, which are usually limited by the hand-crafted manner in specific application. 
Benefit from the success of CNNs on large-scale image repositories such as ImageNet \cite{krizhevsky2012imagenet}, deep neural networks are being applied to process the 3D format data.
As an irregular format of 3D data, point clouds can be transformed into other kinds of regularized format, such as the 3D voxel \cite{Zhizhong2016b,han2017boscc,han2018deep} or the rendered view \cite{han2018seqviews2seqlabels,han20193d2seqviews,han2019y2seq2seq,han2019view,han20193dviewgraph,han2019parts4feature}.
The voxelization of point cloud is a feasible choice, which first converts the point cloud into voxels, and then applies 3D CNNs.
3D ShapeNets \cite{wu20153d} and VoxNet \cite{maturana2015voxnet} represent each voxel with a binary value which indicates the occupied of the location in the space.
However, the performance is largely limited by the resolution loss and the rapid growth of computational complexity.
The inherent sparsity of 3D shapes makes it hard to make full use of the storage of input data, where the hollow inside 3D shapes is often meaningless.
Some improvements \cite{li2016fpnn} have been proposed to alleviate the data sparsity of the volumetric representation.
However, it is still nontrivial to deal with large point clouds with high resolution.

\textbf{Feature Learning from Point Clouds.} PointNet \cite{qi2017pointnet} is a pioneering work which directly adopts point sets as input and obtains convincing performances.
A concise strategy is adopted in PointNet by computing the feature for each point individually and then aggregating these features into a global representation with max-pooling.
However, PointNet is largly limited in capturing the contextual information of local regions.
To address this problem, many recent studies attempt to capture local region contexts.
Specifically, local region contexts can be divided into two categories, which are \emph{intra-region context} and \emph{inter-region context}, respectively.
On the one hand, some studies capture the intra-region context by building graph inside multi-scale local regions.
PointNet++ \cite{qi2017pointnet++} uses sampling and grouping operations for extracting features from several clusters hierarchically to capture the context of each cluster.
 Point2Sequence \cite{liu2018point2sequence} extracts the feature of local regions by a sequence to sequence model with an attention mechanism.
On the other hand, some studies \cite{li2018pointcnn,wang2018dynamic,xu2018spidercnn,wang2017cnn,komarichev2019cnn,hu2019render4completion,wen2020cvpr} investigate CNN-like operations to aggregate neighbors of a given point by building kNN graph inside the single-scale local region.
On the other hand, some studies encode the inter-region context with indexing structures.
KC-Net \cite{shen2018mining} employs a kernel correlation layer and a graph pooling layer for capturing the local structure of point clouds.
ShapeContextNet \cite{xie2018attentional} extends the 2D Shape context \cite{belongie2001shape} to 3D, which divides the local region of a given point into bins and updates the point feature with the aggregation of these bin features.
 KD-Net \cite{klokov2017escape} and OctNet \cite{riegler2017octnet} first divide the input point cloud into leaves,  and then hierarchically extracts features from leaves to the root.
Point2SpatialCapsule \cite{wen2019point2spatialcapsule} integrates capsules to explore the local structures of point clouds, which employs a multi-scale shuffling to increase the diversity of local region features and applies a clustering operation to capture the spatial information of local regions in the feature space.
This complicated procedure significantly differentiates Point2SpatialCapsule from ours.
In general, it is hard for current methods to simultaneously capture the contextual information inside and among multi-scale local regions, which limits the expressiveness of learned representations of point clouds.

 \begin{figure*}[tp]
\centering
\includegraphics[height=5.5cm,width=15cm]{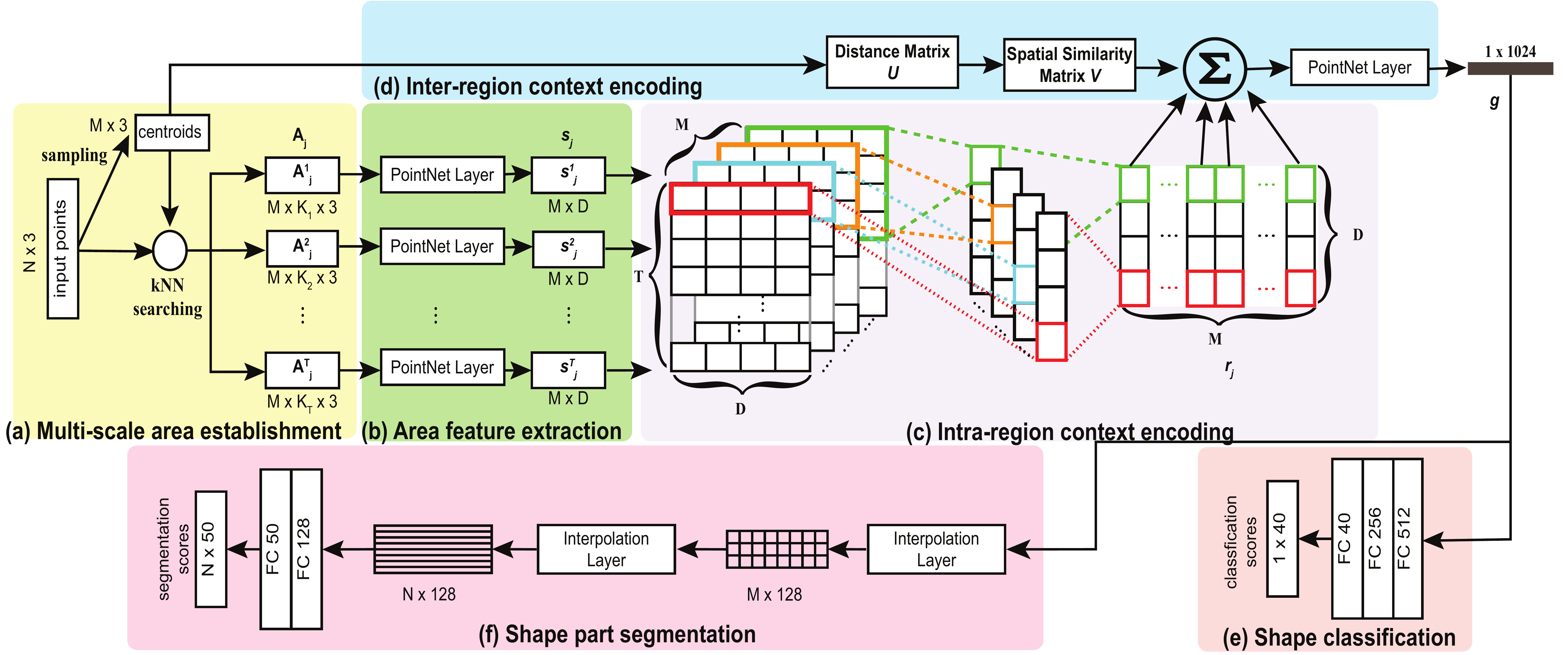}
\caption{\textbf{Our LRC-Net architecture.}\quad In (a), LRC-Net first establishes multi-scale areas inside each local region by sampling and searching layers. Then, PointNet layer is employed to extract the feature of each scale area in (b). Subsequently, the feature of each local region is extracted for intra-region context encoding in (c), which captures the geometric correlation inside the local region. Simultaneously, the spatial similarity measure is calculated for inter-region context encoding in (d), which enhances the spatial relationships among local regions. Finally, the global feature of the point cloud is obtained by aggregating the features of local regions. The learned global feature can be used in shape classification and shape segmentation applications as shown in (e)(f).}
\label{fig:network_arichitecture}
\end{figure*}

\section{The LRC-Net Model}
Figure \ref{fig:network_arichitecture} shows the architecture of LRC-Net, which is composed of six parts: multi-scale area establishment, area feature extraction, intra-region context encoding, inter-region context encoding, shape classification and shape segmentation, respectively. LRC-Net adopts a point cloud $\mathbf{P} = \{ p_i \in \mathbb{R}^3, i=1,2, \cdots,N \}$ as input which is composed of 3D point coordinates x, y and z.
Firstly, a subset with $M$ points, denoted by $\mathbf{P}^{'} = \{ p'_j \in \mathbb{R}^3, j=1,2, \cdots, M \}$, is selected from the input point cloud $\mathbf{P}$ to act as the centroids of local regions $\{ \mathbf{R}_j, j=1, 2,\cdots ,M \}$.
Based on the selected centeroids $\mathbf{P}^{'}$, $T$ different scale areas $\mathbf{A}_j = \{ \mathbf{A}^t_j, t=1,2,\cdots,T\}$ are established in each local region $\mathbf{R}_j$ centered at $p'_j$, where $\{K_t, t=1,2,\cdots, T\}$ points are contained in the each scale area, respectively.
Then, a $D$-dimensional feature $\mathbf{s}^{t}_j$ is extracted from each scale area $\mathbf{A}^{t}_j$.
By stacking $\bm{s}^t_j$, a $T \times D$ feature matrix $\bm{s}_j = \{ \bm{s}^t_j, t = 1,2,\cdots ,T \}$ is formed for each local region $\mathbf{R}_j$ , and further aggregated into a $D$-dimensional feature $\bm{r}_j$ by the intra-region encoding module (see Figure \ref{fig:network_arichitecture}(c)).
Meanwhile, another module of calculating the spatial similarity in the 3D space is applied to capture the inter-region context among local regions (see Figure \ref{fig:network_arichitecture}(d)).
Finally,  a 1024-dimensional feature $\bm{g}$ of the whole input point cloud $\mathbf{P}$ is aggregated from the feature of $M$ local regions, which integrates the extracted intra-region and inter-region context features.
The learned global feature $\bm{g}$ can be applied to shape classification and shape segmentation applications.

\subsection{Multi-scale Area Establishment}
Three key layers are engaged in our structure to establish the multi-scale areas around each sampled point, including sampling layer, searching layer and grouping layer.
The sampling layer uniformly selects $M$ points from the input point cloud $\mathbf{P}$ as the centroids of local regions.
Around each sampled centroid, the searching layer continuously finds $[K_1, \cdots, K_t, \cdots ,K_T]$ nearest points to build the indexing relationship between points.
According to the indexes in the searching layer, the grouping layer groups multi-scale areas $\{ \mathbf{A}^t_j, t=1,2,\cdots,T\}$ inside each local region $\mathbf{R}_j$.

In the sampling layer, farthest point sampling (FPS) is adopted to select $M (M < N)$ points $\mathbf{P}^{'}$ which defines the centroids of local regions.
In the sampling process, the new sampled point $p'_j$ is always the farthest one from previously selected points $\{ p'_1, p'_2, \cdots, p'_{j-1}\}$.
Compared with other sampling methods, such as random sampling, FPS can achieve a more uniform coverage of the entire point cloud with the same number of sampled points.

To build the multi-scale areas $\mathbf{A}_j$, the \emph{k}-nearest neighbors (kNN) algorithm is applied to search the neighbors of a given point based on the Euclidean distance between points.
Another alternative method is the ball query \cite{qi2017pointnet++} which selects all points within a given radius around a point.
Compared with the ball query, kNN can guarantee the information inside local regions and is robust to the input point cloud with different sparsity.
\subsection{Area Feature Extraction}
 As shown in Figure \ref{fig:network_arichitecture}, a concise and effective PointNet layer is employed in LRC-Net to extract the feature for each scale area.
 The PointNet layer is composed of two key parts: a Multi-Layer-Perceptron (MLP) layer and a max-pooling layer, respectively.
The MLP layer individually abstracts the coordinates of points in each area $\mathbf{A}^t_j$ into the feature space, and then these features are aggregated into a $D$-dimensional feature $\bm{s}^t_j$ by the max pooling layer.
So far, a feature map of $T$ different scale areas $\{\mathbf{A}_j,j=1,2,\cdots,M\}$ with the size of $M \times T \times D$ is acquired after the PointNet layer.

Following previous studies \cite{li2018so,qi2017pointnet++},  the relative coordinates are adopted in LRC-Net.
Before feeding points inside each local region $\mathbf{R}_j$ into the PointNet layer, a relative coordinate system of the centroid $p'_j$  is built by a simple operation: $p_l = p_l - p'_j$, where $l$ is the index of points in the local region $\mathbf{R}_j$.
Different from absolute coordinates, the relative coordinates are determined by the relative positional relationship between points.
Therefore, by using relative coordinates, the learned feature of local regions can be invariant to transformations such as rotation and translation.

\subsection{Intra-region Context Encoding}
In order to capture the fine-grained contextual information between multi-scale areas inside local regions, variable-size convolution filters are employed in the architecture.
Inspired by capturing the correlation of different words in the natural language processing tasks \cite{kim2014convolutional}, the intra-region correlation of multi-scale areas is also important in the feature learning of point clouds.
Different from most existing methods that only encode the correlation of fixed scale of areas, we consider capturing the correlation among multiple scales from $1$ to $T$.
As depicted in Figure \ref{fig:network_arichitecture}, given the features $\{\bm{s}^t_j, t=1,2,\cdots,T\}$ of multi-scale areas in a local region $\mathbf{R}_j$ from the area feature extraction module, we first represent these features in a $T \times D$ feature map by
\begin{equation}\label{equation:concat}
 \bm{S}^{1:T}_j = \bm{s}^1_j \oplus \bm{s}^2_j \oplus \cdots \oplus \bm{s}^T_j ,
\end{equation}
where $\oplus$ is the concatenation operator.
In general,  let $\bm{S}^{a:a+b}_j$ refer to the concatenation of features $\bm{s}^a_j, \bm{s}^{a+1}_j, \cdots, \bm{s}^{a+b-1}_j$.
A convolution operation involves a filter $\bm{w} \in \mathbb{R}^{hD}$, which is applied to a window of $h$ scale features to produce a new feature.
For example, a feature $c_k$ is generated from a window of features $\bm{S}^{a:a+h-1}_j$ by
\begin{equation}\label{equation:convolve}
  c_k = f(\bm{w} \cdot \bm{S}^{a:a+h-1}_j + b).
\end{equation}
Here $b \in \mathbb{R}$ is a bias term and $f$ is a non-linear function such as ReLU \cite{nair2010rectified}.
As the intermediate step shown in Figure \ref{fig:network_arichitecture}, this filter is applied to each possible window of scales in the features $\bm{S}^{1:h}_j, \bm{S}^{2:h+1}_j, \cdots , \bm{S}^{T-h+1:T}_j$ to produce a feature vector
\begin{equation}\label{equation:featuremap}
  \bm{c} = [c_1, c_2, \cdots, c_{T-h+1}],
\end{equation}
 with length of $T-h+1$. 
Then, we apply max pooling operation over the feature vector, which extracts the maximum value from feature vector $\bm{c}$ by
\begin{equation}\label{equal:max}
 \hat{\bm{c}} = max \{ \bm{c} \}.
\end{equation}
 Here $\hat{\bm{c}}$ is one element of local region feature $\bm{r}_j$ corresponding to this particular filter.
 So far, we have shown the process of getting one element in $\bm{r}_j$ by a convolution filter with window size $h \times D$.
 In general, there are $T$ kinds of convolution filters in different sizes and $\frac{D}{T}$ filters for each kind of convolution filter.
 Therefore, the output of each input local region $\mathbf{R}_j$ is a $D$-dimensional feature vector $\bm{r}_j$.
\subsection{Inter-region Context Encoding}
To obtain the global feature of point clouds, most existing methods adopt simple pooling layers to aggregate local region features.
However, the inter-region context is largely lost in the pooling process, especially for the spatial distribution information among local regions.
To capture the inter-region spatial context, a greedy strategy is proposed by aggregating the spatial distribution information among local regions in an explicit manner.
Following the intra-region context encoding module, the feature map with the size $M \times D$ of $M$ local regions is obtained.
As shown in Figure \ref{fig:network_arichitecture}, to encode the spatial information of local regions, we explicitly calculate the spatial similarity among local regions based on the coordinates of local region centroids.
Given the coordinate of centroids $\{p'_j, j=1,2,\cdots,M\}$, the $M \times M$ distance matrix $\bm{U}$ is build by
\begin{equation}\label{equation:distance}
 \bm{U} =
 \left[
 \begin{matrix}
 (p'_1-p'_1)^2 & (p'_1-p'_2)^2 & \cdots & (p'_1-p'_M)^2\\
 (p'_2-p'_1)^2 &(p'_2-p'_2)^2 & \cdots & (p'_2-p'_M)^2 \\
 \vdots &\vdots & \ddots &\vdots\\
  (p'_M-p'_1)^2 &(p'_M-p'_2)^2 & \cdots & (p'_M-p'_M)^2
 \end{matrix}
 \right].
\end{equation}
To convert the distance matrix to the similarity space, the spatial similarity matrix $\bm{V}$ is calculated by
\begin{equation}\label{equation:similarity}
\bm{V} = e^{- \gamma \bm{U}}.
\end{equation}
Here $\gamma$ is a parameter which can regulate the effect of the spatial similarity.
Thus, we obtain the spatial similarity among local regions.
To enhance the feature $\bm{r}_j$ of each local region, a greedy weighting strategy is adopted as
\begin{equation}\label{equation:update}
\bm{r}'_j = \sum_{b=1}^{M} \bm{V}_{j,b} \cdot \bm{r}_b,
\end{equation}
where $\bm{r}'_j$ is the enhanced feature vector of $\bm{r}_j$ and $b$ is the index of the column.
In addition, a normalization operation is applied to the enhanced features by
\begin{equation}\label{equation:regularization}
\bm{r}''_j = \frac{\bm{r}'_j}{ \sum_{b=1}^{M} \bm{V}_{j,b}}.
\end{equation}
Here $\bm{r}''_j$ is the final features of local regions after the regularization, which contains the information of spatial distribution among local regions.
In general, it is a greedy stratedgy to compute the spatial similarity between any two local regions.
The greedy strategy aims to enhance the correlation of local regions, which can promote the learning of the global features.
In the subsequent network, a 1024-dimensional global feature $\bm{g}$ of the input point cloud is extracted by another PointNet layer.
The learned global feature $\bm{g}$ can be  applied to various applications, such as shape classification and shape segmentation.

\subsection{Expansion for Shape Segmentation}
The target of shape segmentation is to predict a semantic label for each point in the point cloud.
With the obtained global feature $\bm{g}$, the key is how to acquire the feature for each point.
There are two options, one is to duplicate the global feature with $N$ times as in  \cite{wang2018dynamic}, the other is to perform upsampling by the interpolation layer \cite{qi2017pointnet++}.
In the shape segmentation module, two interpolation layers are equipped in our network, which propagate the features from shape level to point level by upsampling.
The feature propagation $\phi$ between different levels is guided by the inverse distance between $k$-nearest points.
In the interpolation layer, we search $k$ ($k=3$) nearest points for each point in current level from points in previous level.
Therefore, the feature of point $\phi(p)$ in current level is interpolated by the  positional relationship of points between two levels, denoted by
\begin{equation}
\phi(p) = \frac{\sum_{i=1}^{k}{w(p_i)\phi(p_i)}}{\sum_{i=1}^{k}{w(p_i)}},
\label{equation:interpolate}
\end{equation}
where  $w(p_i)=\frac{1}{(p-p_i)^2}$ is the inverse square Euclidean distance between two points, $\phi(p_i)$ is the point feature of $p_i$ and $\{p_i,i=1,2,\cdots,k\}$ are the $k$ nearest points of $p$ in the previous level.
The points in each level are already obtained from the multi-scale area establishment module and the interpolation step can be regard as a reverse process of the abstraction step.
\begin{table*}[tp]
\centering
\scriptsize
\begin{tabular}{p{5cm}|p{1cm}<{\centering}|p{1cm}<{\centering}|p{1cm}<{\centering}|p{1cm}<{\centering}}
\hline
Method &Scales & Points &MN10 & MN40 \\ \hline
 PointNet \cite{qi2017pointnet}   &single &1k   &-   &89.2 \\
 O-CNN \cite{wang2017cnn}       &single & -      &-      &90.6 \\
 MAP-VAE \cite{Han_2019_ICCV}    &single &1k     &94.82  &90.15 \\
 Kd-Net \cite{klokov2017escape} &single &1k &94.0 &91.8 \\
 KC-Net \cite{shen2018mining}  &single &1k &94.4   &91.0 \\
 PointCNN \cite{li2018pointcnn} &single &1k &-    &91.7 \\
 DGCNN \cite{wang2018dynamic}  &single &1k  &-   &92.2 \\
 SO-Net \cite{li2018so}   &single &2k  &94.1 &90.9 \\
 A-CNN \cite{komarichev2019cnn}  &single &1k      &95.5  &92.6 \\
 InterpCNN \cite{mao2019interpolated} & single &1k &-      &93.0 \\
 RS-CNN \cite{liu2019relation}              & single  &1k &-       &93.6 \\
 \hline
  PointNet++ \cite{qi2017pointnet++} &multi &1k &-    &90.7 \\
  L2G-AE \cite{Liu_2019_ACMMM}   &multi &1k &95.37 &90.64 \\
  ShapeContextNet \cite{xie2018attentional} &multi &1k	&-	&90.0 \\
 Point2Sequence  \cite{liu2018point2sequence} &multi &1k &95.3  &92.6 \\
Point2SpatialCapsule \cite{wen2019point2spatialcapsule} &multi &1k &\textbf{95.8} &93.4 \\
 LRC-Net (ours)  &multi &10k &-  &\textbf{94.2} \\
 LRC-Net (ours)  &multi &1k &\textbf{95.8}  &93.1 \\ \hline
\end{tabular}
\caption{The shape classification results (\%) on ModelNet10 and ModelNet40 benchmarks.}
\label{table:classification}
\end{table*}
\begin{table*}[!t]
\resizebox{\textwidth}{!}{
\begin{tabular}{l|c|c|cccccccccccccccccccc}
\hline
\multirow{2}{*}{Method}&
\multicolumn{1}{c|}{\multirow{2}{*}{Scale}}&
	\multicolumn{1}{c|}{\multirow{2}{*}{mean}} &
	\multicolumn{16}{c}{Intersection over Union (IoU)}\\
	& \multicolumn{1}{c|}{} 
    & \multicolumn{1}{c|}{} 
    & \multicolumn{1}{c}{air.}
    & \multicolumn{1}{c}{bag}
    & \multicolumn{1}{c}{cap}
    & \multicolumn{1}{c}{car}
    & \multicolumn{1}{c}{cha.}
    & \multicolumn{1}{c}{ear.}
    & \multicolumn{1}{c}{gui.}
    & \multicolumn{1}{c}{kni.}
    & \multicolumn{1}{c}{lam.}
    & \multicolumn{1}{c}{lap.}
    & \multicolumn{1}{c}{mot.}
    & \multicolumn{1}{c}{mug}
    & \multicolumn{1}{c}{pis.}
    & \multicolumn{1}{c}{roc.}
    & \multicolumn{1}{c}{ska.}
    & \multicolumn{1}{c}{tab.}
    \\ \hline
\# SHAPES & &   &2690 &76 &55 &898 &3758 &69 &787 &392 &1547 &451 &202 &184 &283 &66 &152 &5271 \\
PointNet \cite{qi2017pointnet}			&single &83.7&83.4&78.7&82.5&74.9&89.6&73.0&91.5&85.9&80.8&95.3&65.2&93.0&81.2&57.9&72.8&80.6 \\
Kd-Net	\cite{klokov2017escape}			&single&82.3&80.1&74.6&74.3&70.3&88.6&73.5&90.2&87.2&81.0&94.9&57.4&86.7&78.1&51.8&69.9&80.3 \\
KCNet   	\cite{shen2018mining}		&single&84.7&82.8&81.5&86.4&77.6&90.3&76.8&91.0&87.2&84.5&95.5&69.2&94.4&81.6&60.1&75.2&81.3 \\
DGCNN \cite{wang2018dynamic}&single&85.1 &\textbf{84.2} &83.7 &84.4 &77.1 &90.9 &78.5&91.5 &87.3 &82.9 &96.0 &67.8 &93.3 &82.6 &59.7 &75.5 &82.0	\\
SO-Net 	\cite{li2018so}			&single&84.9&82.8&77.8&88.0&77.3&90.6&73.5&90.7&83.9&82.8&94.8&69.1&94.2&80.9&53.1&72.9&83.0 \\
A-CNN \cite{komarichev2019cnn}                    &single&86.1&84.2&84.0&88.0&79.6&\textbf{91.3}&75.2&91.6&87.1&85.5&95.4&75.3&94.9&82.5&\textbf{67.8}&77.5&83.3 \\
PointCNN \cite{li2018pointcnn}&single &86.1  &84.1 &\textbf{86.5} &86.0 &\textbf{80.8} &90.6 &79.7 &\textbf{92.3} &\textbf{88.4} &85.3 &\textbf{96.1} &\textbf{77.2} &\textbf{95.3} &\textbf{84.2} &64.2 &\textbf{80.0} &83.0
\\
RS-CNN \cite{liu2019relation} &single &\textbf{86.2} &83.5 &84.8 &\textbf{88.8} &79.6 &91.2 &\textbf{81.1} &91.6 &\textbf{88.4} &\textbf{86.0} &96.0 &73.7 &94.1 &83.4 &60.5 &77.7 &\textbf{83.6} \\
\hline
PointNet++  \cite{qi2017pointnet++}		&multi&85.1 &82.4 &79.0 &87.7 &77.3&\textbf{90.8}&71.8&91.0&85.9&83.7&95.3&71.6&94.1&81.3&58.7&\textbf{76.4}&82.6 \\
ShapeContextNet \cite{xie2018attentional}&multi &84.6 &\textbf{83.8} &80.8 &83.5 &\textbf{79.3} &90.5 &69.8 &\textbf{91.7} &86.5 &82.9 &\textbf{96.0}&69.2 &93.8 &82.5 &\textbf{62.9}&74.4 &80.8 \\
Point2Sequence \cite{liu2018point2sequence}	&multi&85.2&82.6&81.8&87.5&77.3&\textbf{90.8}&77.1&91.1&\textbf{86.9}&83.9&95.7&70.8&94.6&79.3&58.1&75.2&\textbf{82.8} \\
Point2SpatialCapsule \cite{wen2019point2spatialcapsule} &multi &\textbf{85.3} &83.5 &83.4 &\textbf{88.5} &77.6 &\textbf{90.8} &79.4 &90.9 &\textbf{86.9} &84.3 &95.4 &\textbf{71.7} & \textbf{95.3} &\textbf{82.6} &60.6 &75.3 &82.5\\
LRC-Net (ours) 	&multi&\textbf{85.3}&82.6&\textbf{85.2}&87.4&79.0&90.7&\textbf{80.2}&91.3&\textbf{86.9}&\textbf{84.5}&95.5&71.4&93.8&79.4&51.7&75.5&82.6 \\ \hline
\end{tabular}}
\caption{The shape segmentation results (\%) on ShapeNet part segmentation dataset.}
\label{table:part_segmentaion}
\end{table*}

\section{Experiments}
In this section, shape classification and shape segmentation applications are adopted to evaluate the performances of the LRC-Net.
In the ablation study, we first investigate how the two main modules affect the performances of LRC-Net in the shape classification task on ModelNet40 \cite{wu20153d}.
Then, we compare our model with several state-of-the-art methods in shape classification on ModelNet10/40 and shape part seqmentation on ShapeNet part dataset \cite{savva2016shrec}.
Finally, some visualizations of the shape segmentation results are also reported.

\subsection{Network Configuration}
In LRC-Net, some network configurations need to be initialized.
According to the input point cloud, we first initialize parameters of the number of sampled points $M = 384$, the number of scales $T = 4$, the number of points in multi-scale areas $K_1 = 16,K_2 = 32,K_3 = 64$ and $K_4 = 128$, the feature dimension $\bm{r}_j$ of each local region $D = 128$.
The rest settings of our model are same as in Figure \ref{fig:network_arichitecture}.
The discussion of some parameters is listed in the Supplementary Material.
In addition, ReLU is used after each fully connected layer with Batch-normalization, and Dropout is also applied with drop ratio 0.4 in the fully connected layers.
In the experiment, we train our network on a NVIDIA GTX 1,080Ti GPU using ADAM optimizer with initial learning rate 0.001, batch size of 16 and batch normalization rate 0.5.
The learning rate and batch normalization rate are decreased by 0.3 and 0.5 for every 20 epochs, respectively.

\section{Parameters Setting}
All the experiments in this section are evaluated on ModelNet40. which contains 40 categories and 12,311 CAD shapes with 9,843 shapes for training and 2,468 shapes for testing.
And the results listed in tables are the instance accuracies.
For each 3D shape, we adopt the point cloud with 1,024 points which are uniformly sampled from the corresponding mesh faces as input.

\begin{table}[htp]
\begin{center}
\begin{tabular}{ccccccc}
\hline
$\gamma$ &0  &1 &${10}^2$  &${10}^4$  &${10}^5$   \\ \hline
Accuracy (\%) &91.33	&91.61 &92.02	&\textbf{93.07}	&92.54  \\ \hline
\end{tabular}
\end{center}
\caption{The effect of the convergence factor $\gamma$ on ModelNet40.}
\label{table:convergence}
\end{table}

In the module of spatial distribution information encoding, $\gamma$ is an important parameter which influences the performance of the whole model.
The results of several settings of $\gamma$ are shown in the Table \ref{table:convergence}.
The best instance accuracy $93.07\%$ is reached at $\gamma = 10^{4}$ which maximizes the effect of the spatial information encoding module.
In particular, $\gamma = 0$ represents a simple summation of the local region features, which will result in the discriminative ability loss of local region features.
From the results, we can see that the spatial information encoding module can promote the global representation learning of point cloud.

\begin{table}[htp]
\begin{center}
\begin{tabular}{ccccc}
\hline
$M$ &128 &256 &384 &512    \\ \hline
Accuracy (\%) &92.22 &92.34 &\textbf{93.07} &92.42	\\ \hline
\end{tabular}
\end{center}
\caption{The effect of the sampled points $M$ on ModelNet40.}
\label{table:sample}
\end{table}

To explore the effect of the sampled points $M$, we keep the setting $\gamma = 10^{4}$ and vary $M$ from 128 to 512 as shown in Table \ref{table:sample}.
The number of the sample points influences the local regions which are visible to the network in the training process.
$M=384$ can obtain a better coverage of all training point clouds, where the input information is balanced between insufficiency and redundancy.

\begin{table}[htp!]
\begin{center}
\begin{tabular}{cccccc}
\hline
$T$ &1 &2 &3 &4 &5    \\ \hline
Accuracy (\%) &92.26 &92.42 &92.54 &\textbf{93.07} &92.18	\\ \hline
\end{tabular}
\end{center}
\caption{The effect of the number of scale areas $S$  in each local region on ModelNet40.}
\label{table:scale}
\end{table}

Moreover, we also discuss the impact of the number of  scale areas $T$ in each local region.
In  the implementation, we keep the number of points 128 in each local region and range $T$ from 1 to 5.
The number of points in each scale is a power of 2 and varies in $[8,16,32,64,128]$.
Specifically, $T=3$ indicates that there are $[32,64,128]$ points in the scale areas.
And similarly, $T=2$ represents there are $[64, 128]$ points in the two scale areas respectively.
In terms of  results in Table \ref{table:scale}, LRC-Net reaches the best performance when the scale areas number is 4.
In practice, the number of scale areas is largely determined by the properties of the input point cloud, especially the sparsity of points.
Therefore, when the number of scale areas in each local region is 4, it is more suitable for our model.
\begin{table}[htp!]
\begin{center}
\begin{tabular}{ccccc}
\hline
$h$ &1 &2 &3 &4    \\ \hline
Accuracy (\%) &92.38 &92.42 &92.67 &\textbf{93.07}	\\ \hline
\end{tabular}
\end{center}
\caption{The effect of the kind of filers $h$ on ModelNet40 in the variable-size convolution module.}
\label{table:filter}
\end{table}

With the number of scale areas to be 4, we change the kind of filters from 1 to 4 in the variable-size convolution module.
In Table \ref{table:filter}, $h=1$ represents only one type of convolutional filter $1 \times D$, and similarly, $h=2$ represents two kind of filters $1 \times D, 2 \times D$.
The experiment results show that the module of variable-size convolution is effective in aggregating the multi-scale area features.

\subsection{Ablation Study}
In the following, we show the effects of the two main modules: the intra-region context encoding and the inter-region context encoding, respectively.
In Table \ref{table:convolution}, we show the performances of LRC-Net with and without the intra-region context encoding module.
Specifically, when we remove the intra-region context encoding, there are three widely used ways to aggregate the features of multi-scale areas by mean pooling (Mean), max pooling (Max) and concatenating (Con), respectively.
\begin{table}[htp]
\begin{center}
\begin{tabular}{ccccc}
\hline
Metric &All &Mean &Max &Con   \\ \hline
Accuracy (\%) &\textbf{93.07} &92.50 &92.38 &92.30	\\ \hline
\end{tabular}
\end{center}
\caption{The effect of intra-region context encoding module in LRC-Net on ModelNet40.}
\label{table:convolution}
\end{table}

 \begin{figure*}[htp]
\centering
\includegraphics[height=5cm,width=\textwidth]{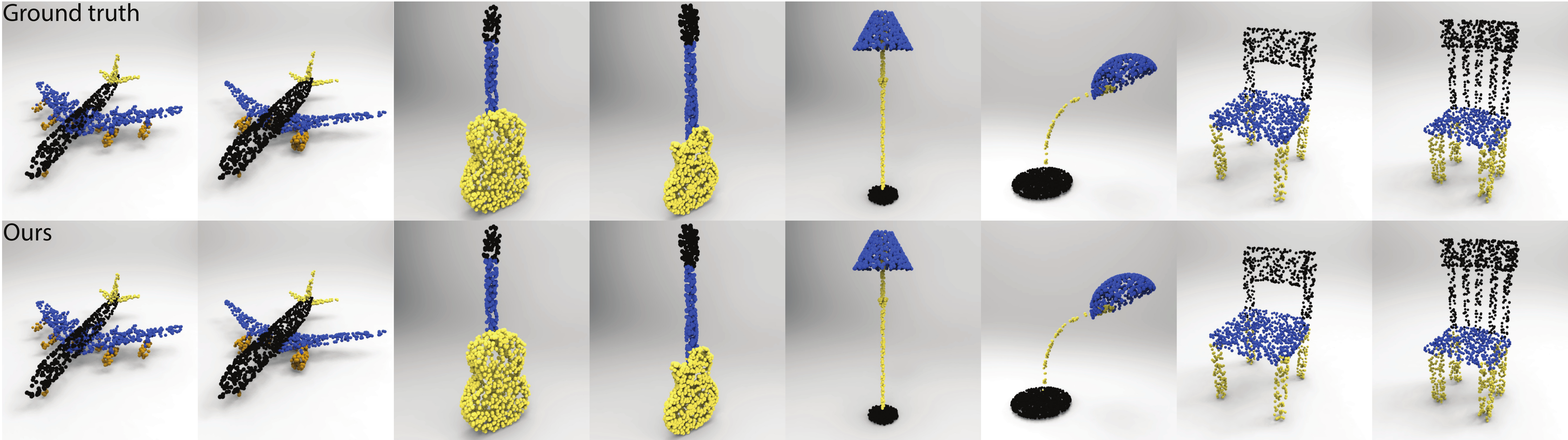}
\caption{Visualization of some shape segmentation results. The top row the is ground truth point clouds, and the bottom row  is our predicted results, where parts with the same color belong to the same class.}
\label{fig:part_seg}
\end{figure*}

These results show that the intra-region context encoding module can promote the discriminative ability of the learned point cloud features.
Similarly, we also evaluate the role of the inter-region context encoding module.
As depicted in Table \ref{table:spatial},  we then list the results with (Y) and without (N) the inter-region context encoding module.
In addition, we also show the influence of the max pooling operation (max) or the mean pooling operation (mean) in the PointNet layer which extracts the global representation $\bm{g}$ as shown in Figure \ref{fig:network_arichitecture}.
Therefore, there are four alternative combinations, Y(max), N(max), Y(mean) and N(mean), respectively.
The results suggests that the inter-region context encoding module is effective in improving the learning of global features  by capturing the spatial context among local regions.
According to above comparisons,  the two modules in LRC-Net are effective in encoding local region contexts.
\begin{table}[htp]
\begin{center}
\begin{tabular}{cccccc}
\hline
Metric &Y(max) &N(max) &Y(mean) &N(mean) &N(sum)   \\ \hline
Accuracy (\%) &\textbf{93.07} &92.26 &91.57 &91.37 &91.41	\\ \hline
\end{tabular}
\end{center}
\caption{The effect of inter-region context encoding module in LRC-Net on ModelNet40.}
\label{table:spatial}
\end{table}

\subsection{Shape Classification}
The performances of LRC-Net are evaluated on both ModelNet10 (MN10) and ModelNet40 (MN40) 3D shape classification benchmarks.
In detail, MN40 contains 40 categories and 12,311 CAD shapes with 9,843 shapes for training and 2,468 shapes for testing.
And MN10 is a subset of MN40 with 4,899 CAD shapes, including 3,991  shapes for training and 908 shapes for testing.
Table \ref{table:classification} compares LRC-Net with several state-of-the-art methods in terms of instance accuracy on MN10 and MN40, respectively.
As shown in the Table \ref{table:classification}, all methods can be divided into two categories: single-scale based methods \cite{li2018pointcnn,komarichev2019cnn} and multi-scale based methods \cite{qi2017pointnet++,liu2018point2sequence}.
LRC-Net has greatly improved the baseline of PointNet++ \cite{qi2017pointnet++} on both ModelNet10 and ModelNet40. 
And LRC-Net achieves the same results with Point2SpatialCapsule \cite{wen2019point2spatialcapsule} on ModelNet10 and reaches comparable results with Point2SpatialCapsule on ModelNet40. 
Point2SpatialCapsule benefits from its network structures such as dynamic routing for clustering and point cloud reconstruction, which aims to increase the network capability. 
However, the two newly added modules (i.e. clustering and point cloud reconstruction) increase both the model size and the computational cost of Point2SpatialCapsule during network learning. 
This makes Point2SpatialCapsule more complicated than LRC-Net in term of the network architecture.
The best accuracy $94.2\%$ is achieved with 10,000 points as input, where the higher resolution point cloud can provide more local details than the sparse input with 1,024 points.
Experimental results show that LRC-Net can effectively enhance the representation learning of point clouds from multi-scale local regions by capturing the contextual information inside and among local regions.

\begin{table}[htp]
  \begin{center}
  \begin{tabular}{lccc}
  \hline
  Method &Model size (MB) &Time (MS) & Accuracy (\%)   \\ \hline
  PointNet (vanilla) \cite{qi2017pointnet}       &9.4   &6.8 &87.1	\\
  PointNet \cite{qi2017pointnet}                  &40 &16.6  &89.2\\
  PointNet++ (SSG) \cite{qi2017pointnet++}       &8.7 &82.4 &- \\
  PointNet++ (MSG) \cite{qi2017pointnet++}          &12 &163.2 &90.7 \\
  PointCNN \cite{li2018pointcnn}       &94 &117.0 &92.3   \\
  LRC-Net (ours)                                   &18 &115.8 &93.1  \\ \hline
  \end{tabular}
  \end{center}
  \caption{Complexity, forward time, and accuracy on ModelNet40 of different models.}
  \label{table:scene}
  \end{table}
In addition, to show the network complexity of LRC-Net intuitively, we make a statistics of model size and space cost of some point cloud based methods.
We follow PointNet++ to evaluate the time and space cost of several point cloud based methods as shown in Table 9. 
We record forward time under the same conditions with a batch size 8 using TensorFlow 1.0 with a single GTX 1080 Ti. 
Table 9 shows LRC-Net can achieve tradeoff between the model complexity (number of parameters) and computational complexity (forward pass time). 
However, influenced by the setting of multi-scale grouping (MSG), LRC-Net and PointNet++ take longer than other single-scale grouping (SSG) based methods.

\subsection{Shape Segmentation}
To further verify the validity of our model, we also evaluate the performance of LRC-Net in the shape segmentation task.
The shape segmentation branch is implemented as depicted in Figure \ref{fig:network_arichitecture}.
In this task, ShapeNet Part dataset is adopt as the benchmark which contains 16,881 models from 16 categories and is spit into train set, validation set and test set as PointNet++.
There are 2,048 points for each point cloud, where each point belongs to a certain one of 50 part classes.
And the kind of semantic parts in each shape varies from 2 to 5.
There is no overlap of the part classes between shapes in different shape categories.

We employ the mean Intersection over Union (IoU) proposed in \cite{qi2017pointnet} as the evaluation metric for shape segmentation.
For each shape, the IoU is computed between ground-truth and the prediction for each part class in the shape category.
And the average IoUs are calculated in each shape category and overall shapes.
In Table \ref{table:part_segmentaion}, we report the performance of LRC-Net in each category and the mean IoU of all testing shapes.

From Table \ref{table:part_segmentaion}, the performance of LRC-Net is not as good as three latest proposed single-scale based methods including PointCNN \cite{li2018pointcnn}, A-CNN \cite{komarichev2019cnn} and RS-CNN \cite{liu2019relation} that adopt some special strategies in the training process. For example, A-CNN states ``We concatenate the one-hot encoding of the object label to the last feature layer" and PointCNN states ``we perturb point locations with the point shuffling for better generalization", which are different from mainstream approaches like PointNet \cite{qi2017pointnet}. For fair comparison with most of other methods, we do not apply these strategies in our method.

Moreover, compared with other multi-scale based methods \cite{qi2017pointnet++,liu2018point2sequence}, LRC-Net achieves the best mean instance IoU of $85.3\%$  and comparable performances on many shape categories, which shows the effective of enhancing the contextual information inside and among local regions.
In addition, some visualizations of the shape segmentation results are shown in Figure \ref{fig:part_seg}, where our predictions are highly consistent with the ground-truths.
The shape segmentation results qualitatively show the effectiveness of LRC-Net in capturing the contextual information for each point.

\begin{table}[htp]
  \begin{center}
  \begin{tabular}{lcc}
  \hline
  Method &Mean IoU &Overall accuracy   \\ \hline
  PointNet (baseline) \cite{qi2017pointnet}       &20.1   &53.2 	\\
  PointNet \cite{qi2017pointnet}                  &47.6 &78.5 \\
  MS + CU (2) \cite{engelmann2017exploring}       &47.8 &49.7 \\
  G + RCU  \cite{engelmann2017exploring}          &49.7 &81.1 \\
  ShapeContextNet \cite{xie2018attentional}       &52.7 &81.6   \\
  LRC-Net (ours)                                     &52.0 &81.3  \\ \hline
  \end{tabular}
  \end{center}
  \caption{The performance of LRC-Net in the semantic segmentation on S3DIS.}
  \label{table:scene}
  \end{table}
\subsection{Indoor Scene Segmentation}
We evaluate our model on Standford Large-Scale 3D Indoor Spaces Dataset (S3DIS)  \cite{armeni20163d} for the semantic scene segmentation task.
There are 6 indoor areas including 272 rooms of the 3D scan point clouds in the S3DIS dataset.
Each point in one point cloud belongs to one of the 13 categories, e.g. chair, board, ceiling and beam.
We follow the same setting as in PointNet \cite{qi2017pointnet}, where each room is split into blocks and 4,096 points are sampled from each block in the training process.
In the testing process, all the points are used.
We also apply the 6-fold cross validation over the 6 areas and report the average evaluation results.

Similar to shape part segmentation task, the probability distribution over the semantic object classes is generated for each input point.
The quantified comparison results with some existing methods are reported in Table \ref{table:scene}.
LRC-Net outperforms PointNet \cite{qi2017pointnet} and achieves comparable results with ShapeContextNet \cite{xie2018attentional}.

\section{Conclusion}
In this paper, we propose a novel feature learning framework for the understanding of point cloud in the shape classification and shape segmentation.
With the intra-region context encoding module, the LRC-Net effectively learns the correlation between multi-scale areas inside each local region.
To enhance the aggregation of local region features, a greedy strategy enables to encode the inter-region context of point clouds.
We justify that both of these two modules are vital to encode local region contexts, which promote learning discriminative feature for point clouds.

\section{Acknowledgments}
This work was supported by National Key R\&D Program of China (2018YFB0505400).

\bibliography{reference}
\end{document}